\def\year{2018}\relax
\documentclass[letterpaper]{article} 
\usepackage{aaai18}  
\usepackage{times}  
\usepackage{helvet}  
\usepackage{courier}  
\usepackage{url}  
\usepackage{graphicx}  
\usepackage[caption=false]{subfig}
\usepackage{amsmath,amssymb}
\usepackage{algorithm,algorithmic}
\usepackage{color}
\usepackage{multirow}
\DeclareMathOperator*{\argmin}{argmin}
\usepackage[american]{babel}

\begin{document}
%
\title{Action Recognition with Coarse-to-Fine Deep Feature Integration and Asynchronous Fusion}
\author{Weiyao Lin$^1$\thanks{This work is supported in part by National Key Research and Development Program of China (2017YFB1002203), NSFC (61471235, 61571297, 61572316, 61720106001), Shanghai `The Belt and Road' Young Scholar Exchange Grant (17510740100), and Tencent research grant. Corresponding author is Weiyao Lin.}, Yang Mi$^1$, Jianxin Wu$^2$, Ke Lu$^3$, Hongkai Xiong$^1$\\
 $^1$ Department of Electronic Engineering, Shanghai Jiao Tong University, China\\
 $^2$ National Key Laboratory for Novel Software Technology, Nanjing University, China \\
 $^3$ University of Chinese Academy of Sciences, China \\
 \{wylin, deyangmiyang, xionghongkai\}@sjtu.edu.cn, wujx2001@nju.edu.cn, luk@ucas.ac.cn
 }

\maketitle

\begin{abstract}
  Action recognition is an important yet challenging task in computer vision. In this paper, we propose a novel deep-based framework for action recognition, which improves the recognition accuracy by: 1) deriving more precise features for representing actions, and 2) reducing the asynchrony between different information streams. We first introduce a coarse-to-fine network which extracts shared deep features at different action class granularities and progressively integrates them to obtain a more accurate feature representation for input actions. We further introduce an asynchronous fusion network. It fuses information from different streams by asynchronously integrating stream-wise features at different time points, hence better leveraging the complementary information in different streams. Experimental results on action recognition benchmarks demonstrate that our approach achieves the state-of-the-art performance.
\end{abstract}

\section{Introduction\label{section:introduction}}

Action recognition, which aims at identifying the action class label for an input action video, has attracted much attention due to its importance in many applications. Although the recent advances in deep convolutional networks (ConvNets) have brought some improvements on action recognition \cite{c3d,KVMF}, it remains challenging due to the large variation of video scenarios and the interferences from noisy contents unrelated to the video topic.

In this paper, we focus on two key issues for improving the performance over the existing ConvNet frameworks: (1) deriving more precise features to better represent actions, (2) reducing the asynchrony among information streams to better leverage the stream-wise complementary information.



First, good features are crucial to reliable action recognition. Although features automatically learned from ConvNets have shown big improvements in many domains~\cite{liu2016two,ImageNet,song2017end}, they make less progress in action recognition due to the high complexity of video data. Some recent studies attempted to improve the deep feature representation of an action by including additional information sources~\cite{dutaspatio,3stream,3stream2}, selecting spatial-temporal attention parts~\cite{kar2016adascan,visualattention,KVMF}, or incorporating more proper temporal information~\cite{TSN,cherian2017generalized}. However, since most of them focus on learning features to directly describe actions' individual action classes, they have limitations in precisely differentiating the ambiguity among action classes due to the large intra-class variations and subtle inter-class differences of actions.

\begin{figure}
  \centering
  \subfloat[]{\includegraphics[width=8cm,height=4.5cm]{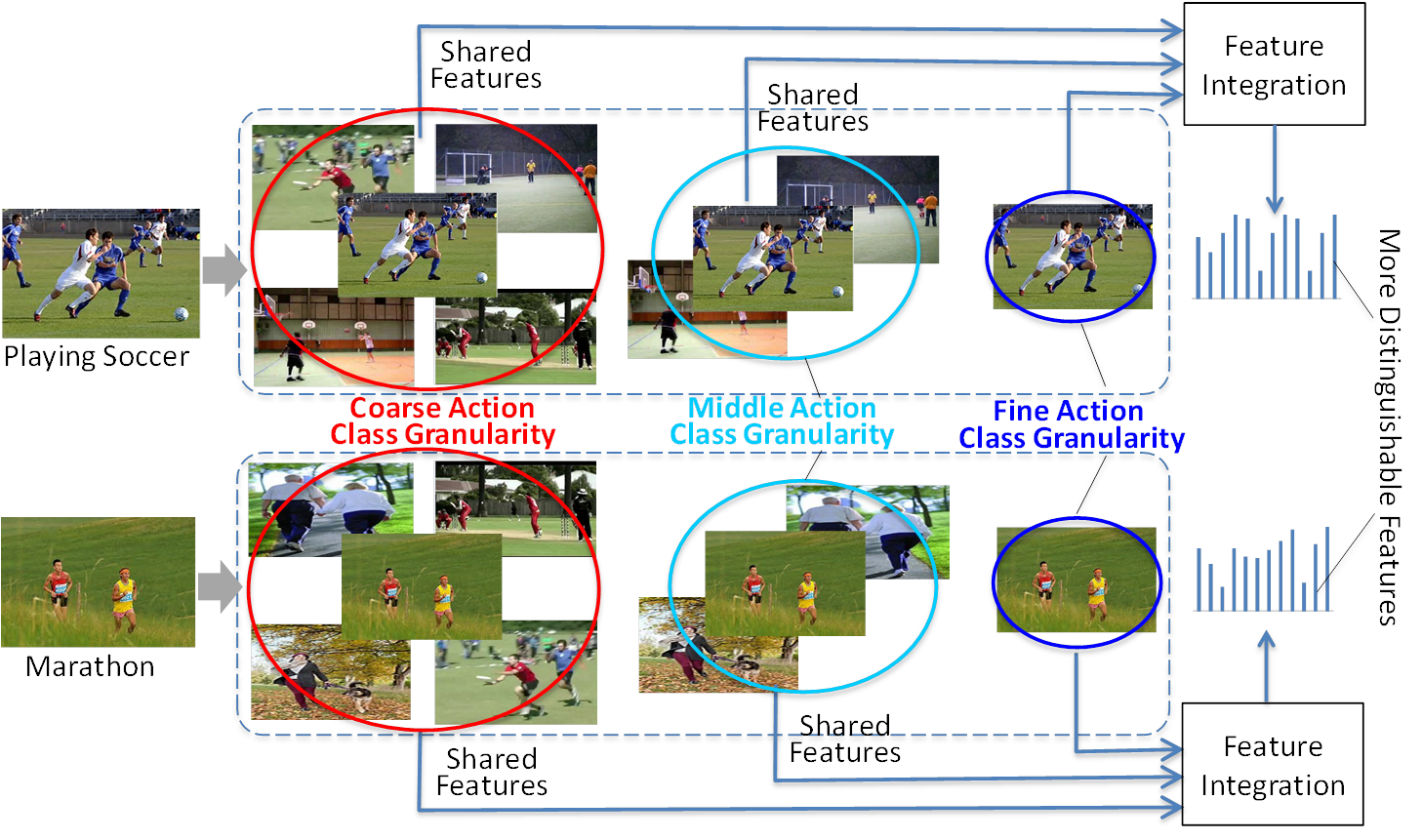} \label{fig:mapping structure_example_b}} \\
  \subfloat[]{\includegraphics[width=0.45\textwidth,height=0.122\textwidth]{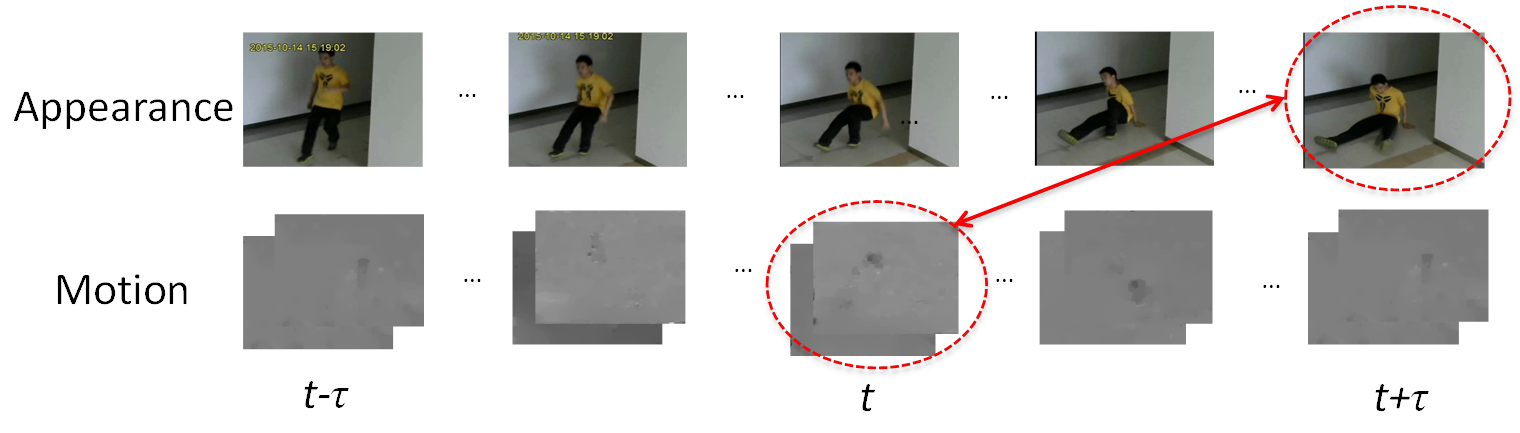} \label{fig:mapping structure_example_c}}
  \caption{(a) Illustration of different action class granularity. (b) Illustration of the asynchronous pattern between streams: The appearance stream is most indicative about ``fall down" after the object has lied down, while the motion stream shows the strongest ``fall down" pattern when the object is in the process of going down. (Best viewed in color)} 
\end{figure}

In this paper, we introduce the idea of \emph{action class granularity} where a coarser action class granularity includes more action classes and a finer action class granularity contains fewer action classes. We argue that features learned for different action class  granularities can provide useful information in discriminating action classes. For example, in Fig.~\ref{fig:mapping structure_example_b}, since the input action clip ``marathon" is visually similar to ``playing soccer", they will be easily confused if directly deriving features to recognize their individual action classes. However, if we relax the recognition requirement from individual action classes to action class groups (i.e., coarser action class granularities), we are able to obtain shared features for representing a set of action classes. These shared features are able to provide more discriminative power as ambiguous action clips may correspond to different groups of action classes in coarser action class granularities (cf. Fig.~\ref{fig:mapping structure_example_b}).

Based on this intuition, we propose a \emph{coarse-to-fine network} which first extracts deep features from different action class granularities, and then progressively integrates them from coarse granularities to fine ones to obtain a precise feature representation for input actions (cf. Fig.~\ref{fig:mapping structure_example_b}). It should be noted that since the action classes in each granularity are only used to derive proper features, they are automatically determined and dynamic for different input video clips.

Second, combining multiple information streams (such as two-stream ConvNets~\cite{baseline}) has shown strong performance and thus has become a mainstream framework in action recognition. However, most existing works only focus on introducing more information streams~\cite{3stream,3stream2} or strengthening the correlation among streams~\cite{TSN,wumultifusion,lstmiccv2017}, while the asynchronous issue among different information streams is less studied.

We argue that many actions have asynchronous patterns in different information streams, which affects the performance of action recognition. For example, Fig.~\ref{fig:mapping structure_example_c} shows two information streams for an action clip ``fall down" (one appearance stream and one motion stream). Apparently, the appearance stream shows the most indicative pattern about ``fall down" after the object has lied down on the floor. Comparatively, the motion stream shows the strongest ``fall down" pattern when the object is in the process of going down. If we simply combine the overall information in both streams or fuse the stream-wise information at the same time point, the indicative patterns appear at different time cannot be fully utilized and the performance is restrained. Therefore, we further introduce an \emph{asynchronous fusion network}, which asynchronously integrates stream-wise features from different time points, hence better leveraging the complementary information in multiple streams.

Overall, our contribution to action recognition are 3 folds:
\begin{enumerate}
 \item We propose a coarse-to-fine network which extracts and integrates deep features from multiple action class granularities to obtain a more precise representation for actions.
 \item We propose an asynchronous fusion network which integrates stream-wise features at different time points for better leveraging the information in multi-streams.
 \item We combine the proposed coarse-to-fine and asynchronous fusion networks into an integrated framework which achieves the state-of-the-art performance.
\end{enumerate}


\section{Related Works\label{section:related_work}}

Action recognition has been studied for years. Early works focus on developing good hand-crafted features for representing actions, such as 3D SIFT~\cite{3dsift} and dense trajectory~\cite{iDT}. The performances for these methods are often restrained due to the limited differentiation capability of hand-crafted features.

With the development of deep ConvNets, many ConvNet-based methods were recently proposed for action recognition, which utilize ConvNets to automatically obtain the feature representation for actions. Ji et al.~\cite{3dcnn2} utilize a 3D ConvNet to recognize actions in video. Simonyan and Zisserman~\cite{baseline} propose a two-stream framework which uses two ConvNets to respectively extract features from two information streams (i.e., appearance and motion) and fuse them for recognition. Based on this framework, recent researches further improve the effectiveness of ConvNet features by including additional information sources~\cite{3stream,3stream2}, selecting spatial-temporal attention parts~\cite{kar2016adascan,visualattention,KVMF}, or incorporating more proper temporal information~\cite{TSN,wumultifusion,cherian2017generalized,DIN}.

Most of the existing works are targeted at learning features for directly describing actions' individual action classes, while the shared characteristics in different action class granularities are less studied. This restrains them from precisely distinguishing the subtle difference among ambiguous actions. Although some methods~\cite{jointattention} obtain different levels of generality by integrating features in multi-ConvNet layers, they still focus on directly representing the individual action classes and do not consider the shared characteristics in different action class granularities.


Besides the derivation of proper features, other researches focus on the proper combination of multiple information streams to boost the action recognition performance~\cite{nips,wumultifusion,twostreamfuse,lstmiccv2017}. For example, Feichtenhofer et al.~\cite{nips} introduce residual connections between information streams to remedy the deficiency of late fusion strategy in the two-stream framework. Wu et al.~\cite{wumultifusion} also improve the fusion efficiency of the two-stream framework by performing both sequence level fusion and video-level fusion over the information streams. However, most of these works fuse stream-wise information that happen simultaneously, which have limitations in handling the longer-term asynchronous pattern among information streams. As will be shown in this paper, the asynchrony among information streams is a non-trivial factor which can bring noticeable performance gains for action recognition.

\section{Overview}
\label{section:overview}

\begin{figure}
  \centering
  \includegraphics[width=0.47\textwidth,height=0.30\textwidth]{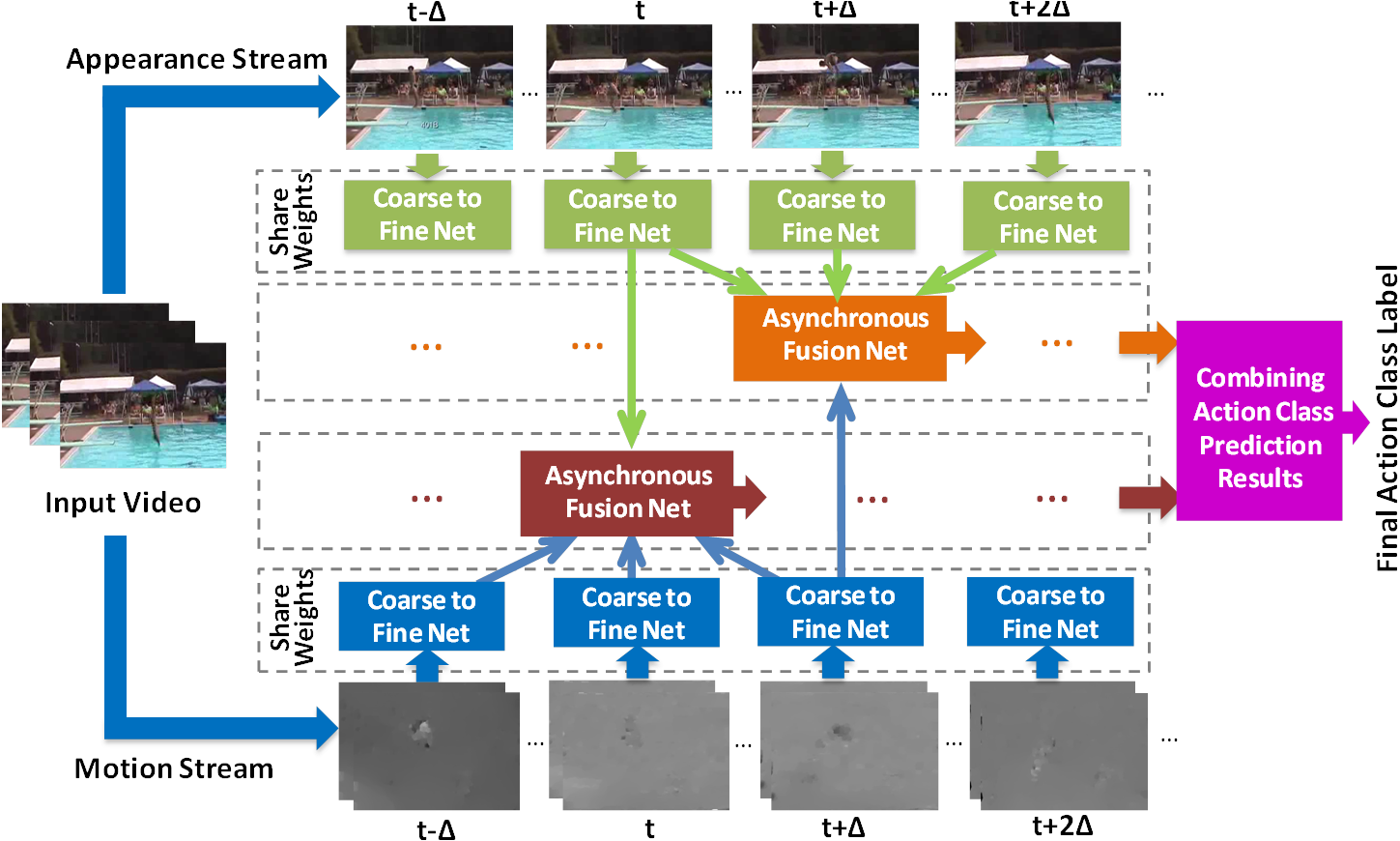}
  \caption{Framework of the approach. The coarse-to-fine network (detailed in Fig.~\ref{fig:coarse-to-fine}) extracts a more precise feature representation for each frame/optical flow stack. These features are then fused by asynchronous fusion networks (detailed in Fig.~\ref{fig:asynchronous}) to obtain action prediction results.}
    \label{fig:framework}
\end{figure}

The framework of our approach is shown in Fig.~\ref{fig:framework}. After obtaining appearance and motion streams from an input video, we first input each spatial frame from the appearance stream and each short-term optical flow stack from the motion stream into a coarse-to-fine network (detailed in Sec.~\ref{section:Coarse-to-fine}), which integrates deep features from multiple action class granularities and creates a more precise feature representation. The extracted features are then fed into asynchronous fusion networks (detailed in Sec.~\ref{section:Asynchronuous}), where each asynchronous fusion network integrates stream-wise features at different time points within a period and obtains an action class prediction result. Finally, action prediction results from different asynchronous fusion networks are combined to decide the final action class of the input video.

Note that the framework of our approach is integrated where the major components in the coarse-to-fine and asynchronous fusion networks can be jointly trained.



\section{Coarse-to-Fine Network\label{section:Coarse-to-fine}}

\begin{figure}
  \centering
  \includegraphics[width=0.47\textwidth,height=0.32\textwidth]{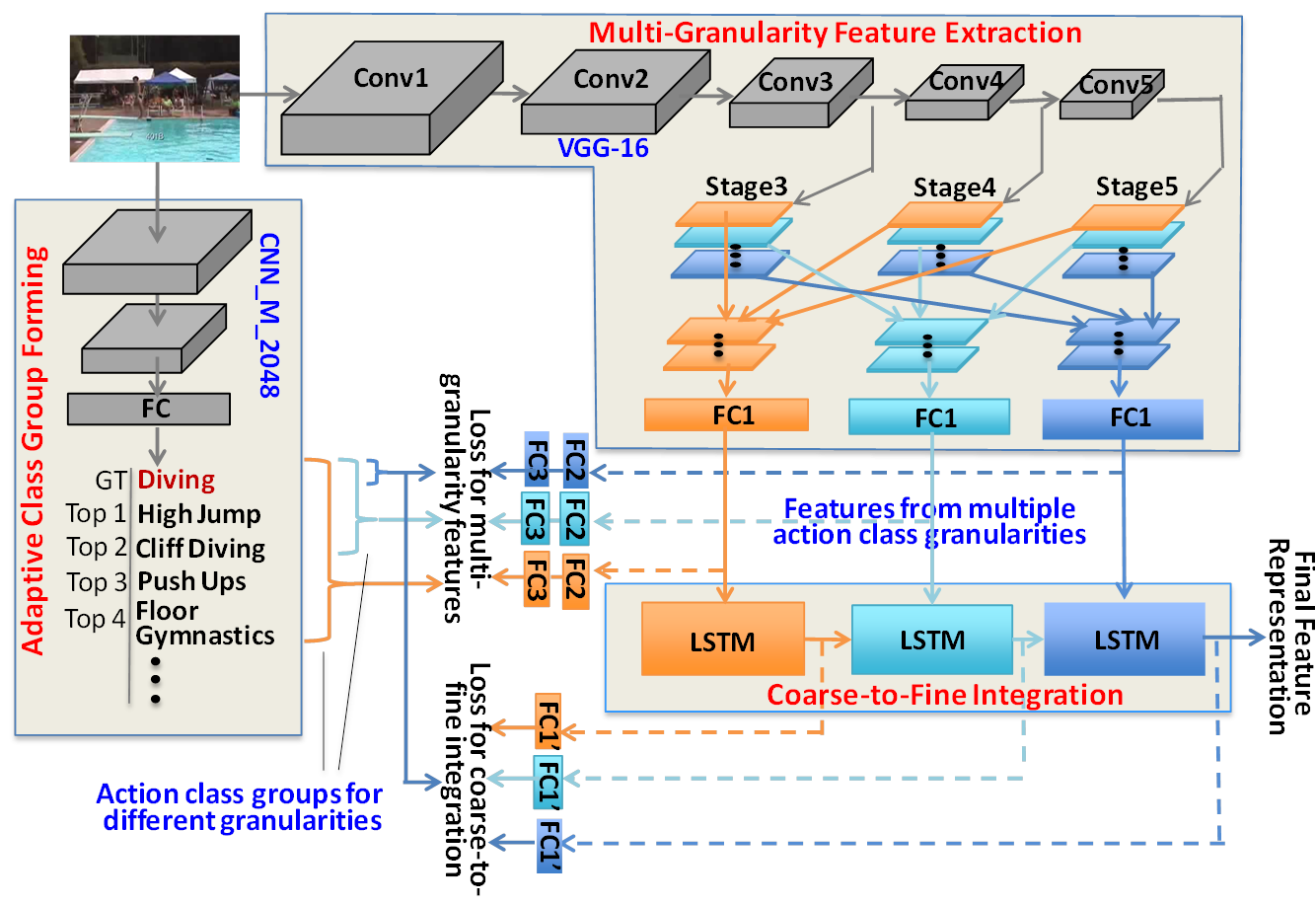}
  \caption{Structure of the coarse-to-fine network.}
    \label{fig:coarse-to-fine}
\end{figure}

The structure of the coarse-to-fine network is shown in Fig.~\ref{fig:coarse-to-fine}. Basically, the network includes three major modules: first, a \emph{multi-granularity feature extraction} module is applied over a ConvNet to extract deep features from different action class granularities. Second, in order to guarantee the extraction of proper features in the feature extraction module, an \emph{adaptive class group forming} module is introduced. This module adaptively forms a suitable action class group for each action class granularity of an input frame/optical flow stack, so as to guide the feature extraction module to create the desired features. Third, a \emph{coarse-to-fine integration} module is connected to the feature extraction module, which progressively integrates features from coarse action class granularities to fine ones and outputs a precise feature representation for the input frame/optical flow stack.

It should be noted that the \emph{adaptive class group forming} module is only used in the training stage, while the \emph{multi-granularity feature extraction} and \emph{coarse-to-fine integration} modules are applied in both training and testing stages.


\subsection{Multi-granularity feature extraction}


The multi-granularity feature extraction module aims to extract deep features from different action class granularities. Since side output layers have shown their effectiveness in encoding multi-scale information in skeleton \& boundary detection~\cite{skeleton,skeleton2}, we borrow them into action recognition and construct three side output flows to extract features in three action class granularities.

Specifically, we derive side output maps from the last convolutional layer in stages $3$, $4$, and $5$ of VGG-$16$ ConvNet (i.e., conv3\_3, conv4\_3, conv5\_3). The side output maps from different stages are then sliced and concatenated into three scale-specific side map groups~\cite{skeleton}, where each side map group corresponds to one action class granularity. In order to ensure output maps from different stages to have the same size, upsampling layers are applied on side output maps before map concatenating. Finally, the scale-specific side map groups are input into a fully connected (FC) layer respectively to obtain features for the three action class granularities (FC$1$ in  Fig.~\ref{fig:coarse-to-fine}). Note that different from the previous side output works~\cite{skeleton,skeleton2}, our approach utilizes an FC layer in the side output flow  to obtain features for describing actions.

\subsection{Adaptive class group forming}

The \emph{adaptive class group forming} module is a key part of the coarse-to-fine network, which aims to form suitable action class groups to guide the feature extraction process in the \emph{multi-granularity feature extraction} module. In this paper, we introduce an additional smaller ConvNet (i.e., CNN\_M\_$2048$~\cite{M2048}) to form action class groups in different granularities.

Specifically, we first use the CNN\_M\_$2048$ ConvNet to predict the action class label of an input frame/optical flow stack, and then use the top $5$, top $3$, and top $1$ action classes in the predicted result to form the action class groups in the three action class granularities, respectively.

Three important issues need to be mentioned about the \emph{adaptive class group forming} module: (1) The \emph{adaptive class group forming} module is only applied in the training stage which helps to construct a reliable multi-granularity feature extracion network. During the testing stage, the feature extraction module will directly output features without the guidance of action class groups. (2) The CNN\_M\_$2048$ ConvNet is pre-trained on the same dataset and is fixed during the training process. We fix the CNN\_M\_$2048$ ConvNet in training in order to create stable action class groups. (3) When forming action class groups, if the groundtruth label of an input frame/optical flow stack is not listed in the top ranked action group in CNN\_M\_$2048$'s prediction result, we will mandatorily include it into the action class group to avoid the feature extraction module deriving irrelevant features to the input frame/optical flow stack.

After action class groups are constructed in the \emph{adaptive class group forming} module, they are used to guide the feature extraction process by a cross-entropy loss~\cite{crossentropy}, which forces the feature extraction module to create shared features that best describe the constructed action class groups in multiple granularities:

\begin{equation}
\begin{aligned}
\mathcal{L}_v(\mathbf{W})=&-\frac{1}{N}\sum_{k=1}^{3}\sum_{n\in \mathbf{G}_k}{{\alpha }_{k}\log{\hat{p}(n|\mathbf{W},k)}}\\
\end{aligned}
\label{equation:equ0}
\end{equation}
where $\mathbf{W}$ is the parameter set for the \emph{multi-granularity feature extraction} module. $N$ is the total number of action classes. $\mathbf{G}_k$ is the constructed action class group for the $k$th action class granularity and ${\alpha }_{k}$ is the weight measuring the relative importance of the $k$th action class granularity. $\hat{p}(n|\mathbf{W},k)$ is the probability for the $n$th action class predicted by the features from $k$th action class granularity. Note that in order to create action prediction results $\hat{p}(n|\mathbf{W},k)$, two additional fully connected layers are added to the feature output layer of the \emph{multi-granularity feature extraction} module in the training stage (FC$2$ \& FC$3$ in Fig.~\ref{fig:coarse-to-fine}).

\begin{figure}
  \centering
  \includegraphics[width=0.47\textwidth,height=0.28\textwidth]{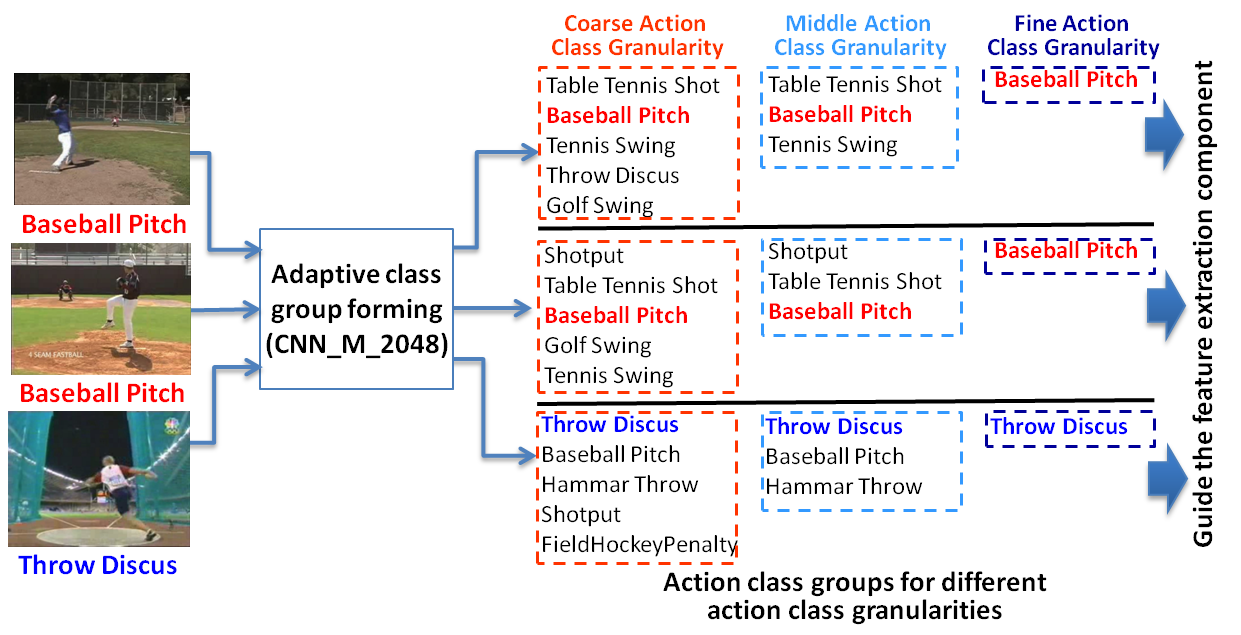}
  \caption{Examples of the adaptively formed action class groups for different inputs (best viewed in color).}
    \label{fig:action-class}
\end{figure}

We argue that by introducing the CNN\_M\_$2048$ ConvNet to form action class groups, we can have three advantages:

\begin{enumerate}
 \item Since the CNN\_M\_$2048$ ConvNet is pre-trained on the same dataset, it has the capability to properly parse the action class contents of an input frame/optical flow stack. Thus, it is able to create informative action class groups which are relatively more similar for same-class inputs and less similar for different-class inputs (cf. the action class groups for the two ``Baseball Pitching" inputs and the ``Throw Discus" input in Fig.~\ref{fig:action-class}). Therefore, when using these action class groups to guide the feature extraction process, we are able to obtain more distinguishable features. Note that the CNN\_M\_$2048$ ConvNet does not need to be perfectly trained. From our experiments, CNN\_M\_$2048$  roughly trained from partial data is already able to create good results (cf. Sec.~\ref{Datasets_experimental settings}).
 \item Since the ground-truth action class of an input frame/optical flow stack is included in each of its action class groups (cf. the red and blue bold action classes in Fig.~\ref{fig:action-class}), features guided by these action class groups are able to capture the characteristics of the input's true action class in different aspects. Therefore, by integrating features from multiple action class granularities (cf. Sec.~\ref{integration}), the feature representation is properly strengthened, which has stronger capability to predict the correct action class for the input sample. 

 \item Moreover, by introducing CNN\_M\_$2048$ ConvNet into our coarse-to-fine network, we are also taking the advantage of properly combining two ConvNets (i.e., CNN\_M\_$2048$ and VGG-$16$) to boost action recognition performances. As will be shown in the experimental results, our approach provides a more proper way to combine ConvNets, which has obviously better performance than only using a single ConvNet or combining ConvNets in simpler ways~\cite{twocnn,twocnn2}.

\end{enumerate}

\subsection{Coarse-to-fine integration}\label{integration}

After obtaining features from multiple action class granularities, we further utilize a \emph{coarse-to-fine integration} module to progressively integrates features from different action class granularities and outputs a precise feature representation. In this paper, we utilize a Long Short Term Memory (LSTM) network to perform coarse-to-fine integration due to its effectiveness in fusing sequential inputs~\cite{LSTM,LSTM2}.



Specifically, we utilize an LSTM model with three units, where each unit takes features $\mathbf{x}_{t}$ from one action class granularity and creates hidden state outputs $\mathbf{h}_{t}$ to influence the next unit (cf. Fig.~\ref{fig:coarse-to-fine}). The hidden state output from the last unit will be the final integrated feature for the input frame/optical flow stack. The entire process is described by: 

\begin{equation}
\begin{aligned}
\mathbf{h}_{1}&={F}_{\mathbf{\Phi}_1}\left ( \mathbf{x}_{1}, 0 \right )
\\\mathbf{h}_{2}&={F}_{\mathbf{\Phi}_2}\left ( \mathbf{x}_{2},\mathbf{h}_{1} \right )
\\\mathbf{h}_{3}&={F}_{\mathbf{\Phi}_3}\left ( \mathbf{x}_{3},\mathbf{h}_{2} \right )
\end{aligned}
\label{equation:equ2}
\end{equation}
where $\mathbf{x}_{t}$ and $\mathbf{h}_{t}$ ($t=1,2,3$) are the input features and hidden state results for $t$th LSTM unit. $\mathbf{\Phi}_t=\{\mathbf{M}_t, \mathbf{b}_t\}$ is the parameter set for $t$th unit and ${F}_{\mathbf{\Phi}_t}$ is the operation of $t$th unit to create hidden state outputs~\cite{LSTM}.  

In the training stage, we utilize the following loss function to train LSTM model to create the desired results.

\begin{equation}
\begin{aligned}
\mathcal{L}_l(&\mathbf{\Phi}_1,\mathbf{\Phi}_2,\mathbf{\Phi}_3)=-\frac{\beta}{N} ( \log{\hat{p}(n_g|\mathbf{\Phi}_1)}\\&+
\log{\hat{p}(n_g|\mathbf{\Phi}_1,\mathbf{\Phi}_2)}+\log{\hat{p}(n_g|\mathbf{\Phi}_1,\mathbf{\Phi}_2,\mathbf{\Phi}_3)}  )
\end{aligned}
\label{equation:equ3}
\end{equation}
where $\mathbf{\Phi}_1,\mathbf{\Phi}_2,\mathbf{\Phi}_3$ are the parameter sets for the three units in LSTM. $\beta$ is the weight measuring the relative importance of the LSTM model. $n_g$ is the ground-truth action class label for an input sample. $N$ is the total number of action classes. $\hat{p}(n_g|\mathbf{\Phi}_1..\mathbf{\Phi}_t)$ is the predicted probability for the ground-truth class from the $t$th unit. Similar to Eq.~\ref{equation:equ0}, in order to create action prediction probability $\hat{p}(n_g|\mathbf{\Phi}_1..\mathbf{\Phi}_t)$, an additional fully connected layer is added to the output of each LSTM unit in the training stage (cf. FC$1'$ in Fig.~\ref{fig:coarse-to-fine}).


\subsection{Loss function for coarse-to-fine network}

The loss function for the coarse-to-fine network is shown by: 

\begin{equation}
\begin{aligned}
\mathcal{L}_{\mathcal{C}}(\mathbf{\Psi}_{\mathcal{C}})=\mathcal{L}_v(\mathbf{W})+\mathcal{L}_l(\mathbf{\Phi}_1,\mathbf{\Phi}_2,\mathbf{\Phi}_3)
\end{aligned}
\label{equation:equ44}
\end{equation}
where $\mathcal{L}_v(\mathbf{W})$ and $\mathcal{L}_l(\mathbf{\Phi}_1,\mathbf{\Phi}_2,\mathbf{\Phi}_3)$ are the losses for the \emph{multi-granularity feature extraction} and \emph{coarse-to-fine integration} modules, respectively. $\mathbf{\Psi_{\mathcal{C}}}=\{\mathbf{W},\mathbf{\Phi}_1,\mathbf{\Phi}_2,\mathbf{\Phi}_3\}$ is the parameter set for the entire coarse-to-fine network.

Note that the coarse-to-fine network can be jointly trained with the asynchronous fusion network in our approach. Therefore, Eq.~\ref{equation:equ44} can be further combined with the loss of the asynchronous fusion network to construct a final loss function for the entire approach, as will be discussed in Sec.~\ref{section:Asynchronuous}.


\section{Asynchronous Fusion Network}\label{section:Asynchronuous}

\begin{figure}
  \centering
  \includegraphics[width=0.43\textwidth,height=0.27\textwidth]{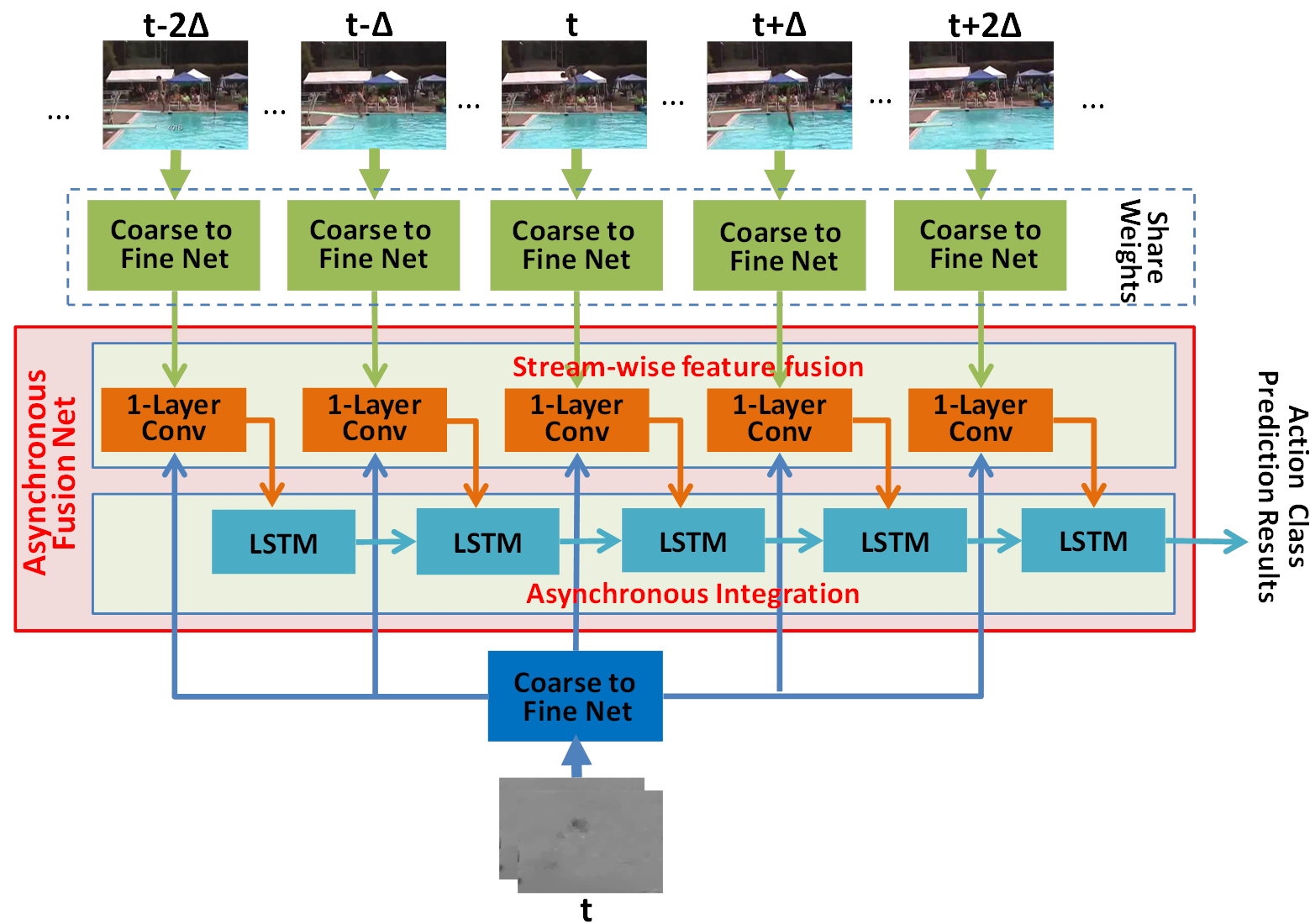}
  \caption{Structure of asynchronous fusion network and its relation with coarse-to-fine networks.} 
    \label{fig:asynchronous}
\end{figure}

The structure of the asynchronous fusion network is shown in Fig.~\ref{fig:asynchronous}. Basically, the asynchronous fusion network aims to fuse an input feature at time $t$ in one stream with multiple input features around $t$ in another stream, so as to leverage the stream-wise complementary information at different time points. It mainly includes two modules: First, the \emph{stream-wise feature fusion} module is used to fuse two input features from different streams. Second, the \emph{asynchronous integration} module is used to integrate the fused outputs over different time and create an action class prediction result for the period of the input features.

\subsection{Stream-wise feature fusion}

Since inputs from different information streams have different characteristics, simply concatenating them may create less effective fusion results. Therefore, we utilize a ConvNet to fuse features from different streams due to its effectiveness in fusing multi-stream inputs~\cite{twostreamfuse}. Since input features are only $1$-dimensional vectors, we simply view them as two $1$-dimensional feature maps and apply a single layer ConvNet with $1\times 1$ kernel to create the fused output.

Note that: (1) In our asynchronous fusion network, an input feature in one stream is fused with $5$ input features from another stream. Therefore, five $1$-layer ConvNets are used to fuse stream-wise features (cf. Fig.~\ref{fig:asynchronous}). (2) Moreover, the five input features to be fused also have $\Delta$ ($\Delta=5$) time intervals to each other. This enables us to capture the longer-term asynchronous patterns between streams.


\subsection{Asynchronous integration}

After obtaining stream-wise fusion results with different time intervals, the \emph{asynchronous integration} module will sequentially integrate them and create an action prediction result for the period of the input features. In this paper, we utilize a five-unit LSTM to perform integration (cf. Fig.~\ref{fig:asynchronous}) since it has good capability in integrating sequential inputs~\cite{LSTM}. 

\subsection{Loss function for the asynchronous fusion network \& the entire framework}

The entire asynchronous fusion network can be trained by:

\begin{equation}
\begin{aligned}
\mathcal{L}_{\mathcal{A}}(\mathbf{\Psi}_\mathcal{A})=-\frac{\gamma}{N} \sum_{t=1}^{T}\log{\hat{p}(n_g|\mathbf{\Phi}_1,..,\mathbf{\Phi}_t,\mathbf{K}_1,..,\mathbf{K}_t)}
\end{aligned}
\label{equation:equ5}
\end{equation}
where $N$ is the total number of action classes. $n_g$ is the ground-truth class label of input video. $T=5$ is the total number of LSTM units and $1$-layer ConvNets. $\mathbf{\Phi}_t$ and $\mathbf{K}_t$ are the parameter sets for the $t$th LSTM unit and $t$th $1$-layer ConvNet, respectively. $\mathbf{\Psi_{\mathcal{A}}}=\{\mathbf{\Phi}_1,...,\mathbf{\Phi}_T,\mathbf{K}_1,...,\mathbf{K}_T\}$ and $\gamma$ are the parameter set and weight for the entire asynchronous fusion network, respectively. $\hat{p}(n_g|\mathbf{\Phi}_1,...,\mathbf{\Phi}_t,\mathbf{K}_1,...,\mathbf{K}_t)$ is the predicted probability for the ground-truth class from the $t$th LSTM unit.


Moreover, the asynchronous fusion network can be jointly trained with the coarse-to-fine network by combining their loss functions. Therefore, the overall framework of our approach can be trained by:

\begin{equation}
\begin{aligned}
(&\mathbf{\Psi}_{\mathcal{C},s_1},\mathbf{\Psi}_{\mathcal{C},s_2},\mathbf{\Psi}_{\mathcal{A}})^*= \\ &\mathop{\argmin}{(\sum_{t=1}^T{\mathcal{L}_{\mathcal{C}}^{t}(\mathbf{\Psi}_{\mathcal{C},s_1}})}+  \mathcal{L}_{\mathcal{C}}(\mathbf{\Psi}_{\mathcal{C},s_2})+\mathcal{L}_{\mathcal{A}}(\mathbf{\Psi}_\mathcal{A}))
\end{aligned}
\label{equation:equ6}
\end{equation}
where $\mathbf{\Psi}_{\mathcal{C},s_1}$, $\mathbf{\Psi}_{\mathcal{C},s_2}$ are the parameter sets of the coarse-to-fine networks for the first and second information streams. $\mathbf{\Psi}_{\mathcal{A}}$ is the parameter set of the asynchronous fusion network. $\mathcal{L}_{\mathcal{C}}(\cdot)$ and $\mathcal{L}_{\mathcal{A}}(\cdot)$ are the loss functions of the coarse-to-fine and asynchronous fusion networks (cf. Eqs.~\ref{equation:equ44} and \ref{equation:equ5}). $T=5$ is the total number of inputs in the first stream (cf. Fig.~\ref{fig:asynchronous}). Note that since the five coarse-to-fine networks in the first stream share weights, we use the same parameter set $\mathbf{\Psi}_{\mathcal{C},s_1}$ to calculate the loss of each input $\mathcal{L}_{\mathcal{C}}^{t}(\mathbf{\Psi}_{\mathcal{C},s_1}), t=1,...,5$.

Besides, it should also be noted that our approach actually requires to construct two independent models, where one model fuses an appearance-stream input with multiple motion-stream inputs, and another model fuses a motion-stream input with multiple appearance-stream inputs. The action prediction results from both models and at different time periods are then combined to decide the final label of an input video (cf. Fig.~\ref{fig:framework}). In this paper, we follow the mainstream two-stream methods~\cite{TSN} to combine action prediction results, which adds the action prediction results from different models \& periods and selects the class with the largest overall prediction score as the final result.

\section{Experimental Results\label{section:experimental evaluation}}

\subsection{Datasets \& experimental settings}\label{Datasets_experimental settings}

{\bf Datasets.} We perform experiments on two benchmark datasets: UCF$101$~\cite{ucf101} and HMDB$51$~\cite{hmdb51}. UCF$101$ dataset is a commonly used dataset for action recognition. It contains $13,320$ video clips in $101$ action classes. HMDB$51$ dataset is a large collection of realistic videos, which contains $6,766$ video clips in $51$ action classes.


{\bf Experimental settings.} We implement our approach on Caffe~\cite{caffe}. The batch size and momentum are set to be $16$ and $0.9$, respectively. The weight parameters for different granularities in the coarse-to-fine network ($\alpha_1,\alpha_2,\alpha_3$ in Eq.~\ref{equation:equ0}) are set to be $0.1$, $0.1$, $1$ respectively. Besides, the weight parameters for the LSTM models in the coarse-to-fine and asynchronous fusion networks ($\beta$ and $\gamma$ in Eqs.~\ref{equation:equ3} and \ref{equation:equ5}) are set as $2$ to let the networks focus more on the reliability on their final outputs.

We use the same method as~\cite{TSN,TDDIDT,transform} to construct optical flow stacks and perform data augmentation. Moreover, the VGG-16 models in both appearance and motion streams are initialized with a pre-trained model from ImageNet~\cite{ImageNet}. When training the entire framework, we set the initial learning rate as $10^{-2}$ and is decreased to its $1/10$ for every $20$K iterations. The maximum iteration is $100$K. Besides, the CNN\_M\_$2048$ ConvNet used to construct action class groups are trained with $1/8$ of the training data and $10$K iterations.

During evaluation, we sample $12$ periods from each video, where each period includes $5$ frames and $5$ corresponding optical flow stacks with temporal distance $\Delta=5$ (cf. Fig.~\ref{fig:asynchronous}). The prediction results from these periods are combined to obtain the final result.


\subsection{Results for the coarse-to-fine network}\label{section:results for coarse to fine}

In order to evaluate the effectiveness of our coarse-to-fine network, we compare six methods: (1) The standard two-stream approach~\cite{baseline} ({\emph{Two-stream baseline}}). (2) Combining $2$ two-stream networks (VGG-$16$ and CNN\_M\_$2048$) by fusing their fully connected layers for recognition~\cite{twocnn,twocnn2} ({\emph{Direct combine ConvNets}}). (3) Delete the {\emph{adaptive class group forming} module and only use the loss function in Eq.~\ref{equation:equ3} to train the coarse-to-fine network ({\emph{CO2FI-no class grouping}}). (4) Delete the coarsest action class granularity and only use the two finer action class granularities in our coarse-to-fine network for recognition ({\emph{CO2FI-two granularities}}). (5) Use three action class granularities in the coarse-to-fine network, but each granularity only contains a single ground-truth action class ({\emph{CO2FI-no coarseness}}). (6) The complete version of our coarse-to-fine network ({\emph{CO2FI-complete}}). 

\begin{table}
\centering
\caption{Results of coarse-to-fine network (split1)}\label{tab:cmcTable1}
\scriptsize{
\label{table1}
\begin{tabular}{|p{0.85cm}|c|*{5}{c|}c}
\hline
{}&\textbf{Methods}& Appearance& Motion& 2-stream \\
\hline
\multirow{6}*{{\tiny{\bf{UCF101}}}}
&Two-stream baseline& {79.2\%}& {84.8\%}& {89.8\%} \\
&Direct combine ConvNets& {80.1\%}& {85.4\%}& {90.6\%} \\
&CO2FI-no class grouping& {79.1\%}& {85.2\%}& {90.0\%} \\
&CO2FI-two granularities& 81.0\%& 86.9\%& 91.7\% \\
&CO2FI-no coarseness& 79.5\%& 85.4\%& 90.4\% \\
&{\bf CO2FI-complete}& {\bf 81.7\%}& {\bf 87.9\%}& {\bf 92.8\%}  \\
\hline
\multirow{2}*{{\tiny{\bf{HMDB51}}}}
&Two-stream baseline& {48.1\%}& {55.4\%}& {58.4\%} \\
&{\bf CO2FI-complete}& {\bf 55.5\%}& {\bf 63.0\%}& {\bf 67.9\%}  \\
\hline
\end{tabular}}
\end{table}

Table~\ref{tab:cmcTable1} compares the action recognition results on split $1$ of UCF$101$ and HMDB$51$ datasets, where the mean classification accuracy for appearance stream, motion stream, and two-streams are listed. Note that in order to delete the effect of the asynchronous fusion network in this experiment, we directly add a softmax layer after the coarse-to-fine network to obtain recognition results. From Table~\ref{tab:cmcTable1}, we can observe:

(1) The performance of the \emph{CO2FI-no class grouping} method is similar to \emph{two-stream baseline} and is obviously lower than the complete version of our approach (\emph{CO2FI-complete}). This implies that without the guidance of the \emph{adaptive class group forming} module, the coarse-to-fine network will construct less precise features and bring few improvements. Besides, the $\emph{Direct combine ConvNets}$ method also achieves less obvious improvements. This further indicates that satisfactory results cannot be easily obtained without a proper way to combine ConvNets.

(2) Comparing the \emph{CO2FI-no coarseness} method with the \emph{CO2FI-two granularities} method, we can see that less noticeable improvements are obtained if each action class granularity only contains one ground-truth action class (\emph{CO2FI-no coarseness}). Comparatively, when each action class granularity includes more action classes, more obvious improvements are achieved with only two action class granularities (\emph{CO2FI-two granularities}). This indicates that the shared characteristics from multiple action classes are the key parts to improve feature representations, and the improvements are restrained if these shared characteristics cannot be obtained (as in \emph{CO2FI-no coarseness}).

(3) The complete version of our coarse-to-fine network (\emph{CO2FI-complete}), which obtains features by including more action class granularities with different coarseness, has the largest improvement over the baseline. This further demonstrates the effectiveness of our approach.




\subsection{Results for the asynchronous fusion network}

Table~\ref{tab:cmcTable2} shows the performance of our asynchronous fusion network. In Table~\ref{tab:cmcTable2}, the upper part shows the results by applying the fusion network on the baseline two-stream ConvNet (i.e., \emph{Baseline+SYN}, \emph{Baseline+ASYN}), and the lower part shows the results by combining our fusion network with the coarse-to-fine network (\emph{CO2FI+ASYN}). Moreover, \emph{SYN} refers to the method that fuses two stream-wise features at the same time point. \emph{ASYN ($\Delta=1$)} and \emph{ASYN ($\Delta=5$)} mean using our asynchronous fusion network to fuse stream-wise features, where the temporal distances of input features being fused are $1$ and $5$ (cf. Fig.~\ref{fig:asynchronous}).

\begin{table}
\centering
\caption{Results of asynchronous fusion network (split 1) 
}\label{tab:cmcTable2}
\scriptsize{
\begin{tabular}{|p{4.5cm}|c|*{5}{c|}c}
\hline
\textbf{Methods}& UCF101& HMDB51 \\
\hline
Two-stream baseline &  89.8\%& 58.4\% \\
Baseline+SYN & 89.7\%& --\\
Baseline+ASYN ($\Delta=1$) & 90.3\% & -- \\
{\bf Baseline+ASYN ($\Delta=5$)}&  {\bf 91.0\%} &{\bf 60.9\%}\\
\hline
CO2FI & 92.8\%& 67.9\%  \\
{\bf CO2FI+ASYN ($\Delta=5$)}& {\bf 93.7\%}& {\bf 69.5\%}  \\
\hline
\end{tabular}}
\end{table}


From the upper part of Table~\ref{tab:cmcTable2}, we can see that simply fusing features at the same time point brings no improvements (\emph{Baseline+SYN}). When we only fuse stream-wise features that are temporally close to each other (\emph{Baseline+ASYN ($\Delta=1$)}), the improvements are still less obvious since the longer-term asynchronous patterns are not properly captured. Comparatively, when fusing stream-wise features with larger temporal distances (\emph{Baseline+ASYN ($\Delta=5$)}), we can obtain more noticeable improvements. This demonstrates that the asynchrony between different information streams indeed affects action recognition performances. Moreover, from the lower part of Table~\ref{tab:cmcTable2}, we can also observe that when combining our asynchronous fusion network with the coarse-to-fine network, we can obtain further improved recognition performances by leveraging both mutli-granularity features and stream-wise complementary information (\emph{CO2FI+ASYN ($\Delta=5$)}).

Fig.~\ref{fig:exp} further shows an example about the effect of the asynchronous fusion network. In Fig.~\ref{fig:exp}, since the two information streams of the ``Highjump" video have asynchronous patterns, they create high prediction scores for the ground-truth action class at different time points (e.g., $t_3$ for appearance stream and $t_2$  for motion stream in Fig.~\ref{fig:exp}). If we simply sum up the prediction scores over time or only consider the stream-wise correlation at the same time, the final recognition result will be confused with other action classes (cf. \emph{Overall score w/o ASYN}). Comparatively, if we consider the asynchrony between streams and allow stream-wise feature fusion at different time, the complementary information between streams can be more properly used, resulting in a correct result (cf. \emph{Overall score with ASYN} in Fig.~\ref{fig:exp}).

\begin{figure}
  \centering
  \includegraphics[width=0.47\textwidth,height=0.18\textwidth]{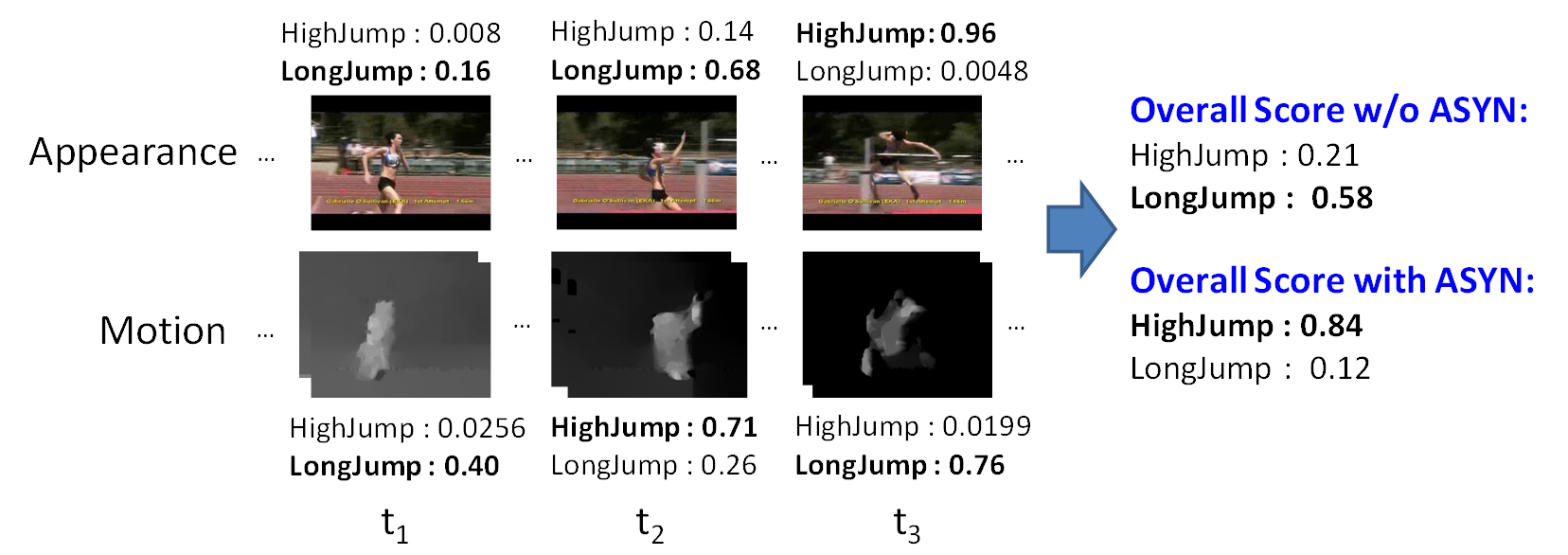}
  \caption{An example of the effect of the asynchronous fusion network: since two streams create high prediction scores for the ground-truth class ``Highjump" at different time points, the recognition result will be easily confused if not considering the stream-wise asynchronous pattern.} 
    \label{fig:exp}
\end{figure}

\subsection{Comparison with the state-of-the-art}

Table~\ref{tab:cmcTable3} compares our approach (\emph{CO2FI+ASYN}) with the state-of-the-art methods. Since many works reported results by performing a late fusion with hand-crafted IDT features~\cite{iDT}, we also show fusion result of our approach (\emph{CO2FI+ASYN+IDT}}). Note that in this experiment, we adopt three training/testing splits on both datasets in order to have a fair comparison with other methods.



From Table~\ref{tab:cmcTable3}, we can see that our approach has better performances than most of the state-of-the-art methods. Specifically, when comparing with the most recent works using ResNet (\emph{ST-ResNet}) or introducing an additional information stream (\emph{ST-VLMPF}), our approach can also obtain similar or better results. This demonstrates the effectiveness of our proposed approach. Note that comparing with \emph{ST-ResNet} and \emph{ST-VLMPF}, we use a relatively short ConvNet (VGG-$16$) and do not introduce additional information streams. It is expected that the performances of our approach can be further improved if using deeper ConvNets such as ResNet or including more information streams. Moreover, our approach fused with IDT (\emph{CO2FI+ASYN+IDT}) also performs better than other IDT-fused methods. This furhter indicates the robustness of our approach in improving performances.



\begin{table}
\centering
\caption{Comparison of different methods (3 splits) 
}\label{tab:cmcTable3}
\scriptsize{
\begin{tabular}{|p{5.3cm}|c|*{5}{c|}c}
\hline
\textbf{Methods}& UCF101& HMDB51 \\
\hline
C3D (3 nets) [Tran \emph{et al.} 2015]&  85.2\%& -- \\
AdaScan [Kar \emph{et al.} 2017] &  89.4\%& 54.9\% \\
TDD+FV [Wang \emph{et al.} 2015]&  90.3\%& 63.2\% \\
GRP [Cherian \emph{et al.} 2017]&  91.9\%& 65.4\%\\
Three-stream sDTD [Shi \emph{et al.} 2017]&  92.2\%& 65.2\%\\
Transformations [Wang \emph{et al.} 2016] &  92.4\%& 62.0\%\\  
Two-Stream Fusion [Feichtenhofer \emph{et al.} 2016]&  92.5\%& 65.4\%\\
KVMF [Zhu \emph{et al.} 2016]&  93.1\%& 63.3\%\\
ST-ResNet [Feichtenhofer \emph{et al.} 2016]&  93.4\%& 66.4\%\\
L$^2$STM [Sun \emph{et al.} 2017]&  93.6\%& 66.2\%\\
ST-VLMPF [Duta \emph{et al.} 2017]&  93.6\%& {\bf 69.5}\%\\
TSN (2 modelities) [Wang \emph{et al.} 2016b] &  94.0\%& 68.5\%\\
{\bf CO2FI + ASYN}& {\bf 94.3\%}& {69.0\%}  \\
\hline
Dynamic Image Networks + IDT [Bilen \emph{et al.} 2016] &  89.1\%& 65.2\% \\
AdaScan + IDT [Kar \emph{et al.} 2017] &  91.3\%& 61.0\% \\
TDD + IDT [Wang \emph{et al.} 2015]&  91.5\%& 65.9\%\\
GRP + IDT [Cherian \emph{et al.} 2017]&  92.3\%& 67.0\%\\
ST-ResNet + IDT [Feichtenhofer \emph{et al.} 2016]&  94.6\%& 70.3\%\\
{\bf CO2FI + ASYN + IDT}& {\bf 95.2\%}& {\bf 72.6\%}  \\
\hline
\end{tabular}}
\end{table}


\section{Conclusion\label{section:conclusion}}

This paper presents a novel framework for action recognition. Our framework consists of two key ingredients: 1) a coarse-to-fine network, which extracts and integrates deep features from multiple action class
granularities to obtain a more precise feature representation for actions; 2) an asynchronous fusion network which integrates stream-wise features at different time points
for better leveraging the information in multiple streams.
Experimental results show that our approach achieves the state-of-the-art performance.


\bibliography{egbib}
\bibliographystyle{aaai}
\end{document}